\documentclass{article}
\usepackage{iclr2027_conference,times}

\usepackage{amsmath,amsfonts,bm}

\def\eqref#1{equation~\ref{#1}}

\def\1{\bm{1}}

\DeclareMathAlphabet{\mathsfit}{\encodingdefault}{\sfdefault}{m}{sl}
\SetMathAlphabet{\mathsfit}{bold}{\encodingdefault}{\sfdefault}{bx}{n}

\usepackage{amsmath}
\usepackage{amssymb}
\usepackage{booktabs}
\usepackage{capt-of}
\usepackage{float}
\usepackage{graphicx}
\usepackage{multirow}
\usepackage[colorlinks=true, linkcolor=blue, citecolor=blue, urlcolor=blue]{hyperref}
\usepackage{url}
\hypersetup{
    pdftitle={VisualErase: Dual-Branch Visual Trajectory Redirection for Robust Concept Erasure in Text-to-Image Diffusion Models},
    pdfauthor={Qianlong Xiang, Miao Zhang, Kun Wang, Yupeng Hu, Junhui Hou, Liqiang Nie}
}

\newcommand{\method}{\textsc{VisualErase}}

\title{VisualErase: Dual-Branch Visual Trajectory Redirection for Robust Concept Erasure in Text-to-Image Diffusion Models}

\author{%
\normalfont\normalsize\bfseries
Qianlong Xiang$^{1,2,3}$, Miao Zhang$^{1}$, Kun Wang$^{4}$, Yupeng Hu$^{5}$, Junhui Hou$^{2}$, Liqiang Nie$^{1}$\\[0.4em]
\normalfont\normalsize
$^{1}$Harbin Institute of Technology (Shenzhen)\quad $^{2}$City University of Hong Kong\\
$^{3}$Shenzhen Loop Area Institute\quad $^{4}$National University of Singapore\quad
$^{5}$Shandong University
}

\iclrfinalcopy

\begin{document}

\maketitle
\lhead{Preprint}
\rhead{}
\cfoot{\thepage}
\renewcommand{\headrulewidth}{0.4pt}

\begin{abstract}
Concept erasure is essential for the safe deployment of text-to-image diffusion models, as they may reproduce harmful, copyrighted, or privacy-sensitive content learned from unconstrained large-scale data.
Existing methods typically erase unwanted concepts while preserving general generation capability by redirecting target-related text-to-image mappings.
However, recent studies show that erased models may still retain visual generative trajectories of target concepts, leaving them vulnerable to adversarial recovery attacks and revealing a fundamental gap between redirecting text-to-image mappings and truly removing visual knowledge.
To bridge this gap, we propose \method{}, a new paradigm that redirects concept-bearing visual generative trajectories toward explicitly defined concept-removed outcomes.
To enable this redirection, we use structure-preserving image editing to construct content-aligned, concept-removed counterparts for source images, providing explicit visual endpoints that retain non-target content.
We then derive a denoising target from each source-to-counterpart pair and use a dual-branch redirection loss to align both text-conditioned and unconditional predictions with this target, since conditional supervision alone does not explicitly constrain generation without textual guidance.
To mitigate the adverse effects of concept erasure on non-target generation, we jointly optimize the redirection loss with a counterpart retention loss that matches denoising predictions from the frozen pretrained model.
Across style, celebrity, and nudity erasure, \method{} limits the maximum attack success rate over seven attacks to 0\%, 8\%, and 0.1\%, respectively, while retaining general generation quality.
These results highlight the importance of visual trajectory redirection for robust concept erasure beyond text-to-image mappings alone.
The project page is \url{https://qianlong0502.github.io/VisualErase-Homepage/}.
\end{abstract}

\section{Introduction}
Text-to-image diffusion models can synthesize diverse, high-quality images from natural-language descriptions \citep{rombach2022high,nichol2021glide,saharia2022photorealistic}.
Their capabilities emerge from training on web-scale image-text corpora, for which exhaustive content filtering remains difficult \citep{schuhmann2022laion}.
Consequently, pretrained models can reproduce copyrighted artistic styles, recognizable identities, and explicit content \citep{jiang2023ai,carlini2023extracting,somepalli2023diffusion,schramowski2023safe}.
Concept erasure aims to address these risks by removing a designated concept from a pretrained model while preserving generation of non-target content. \citep{gandikota2023erasing,kumari2023ablating,zhang2024forget}.

\begin{figure}[t]
    \centering
    \includegraphics[width=\linewidth]{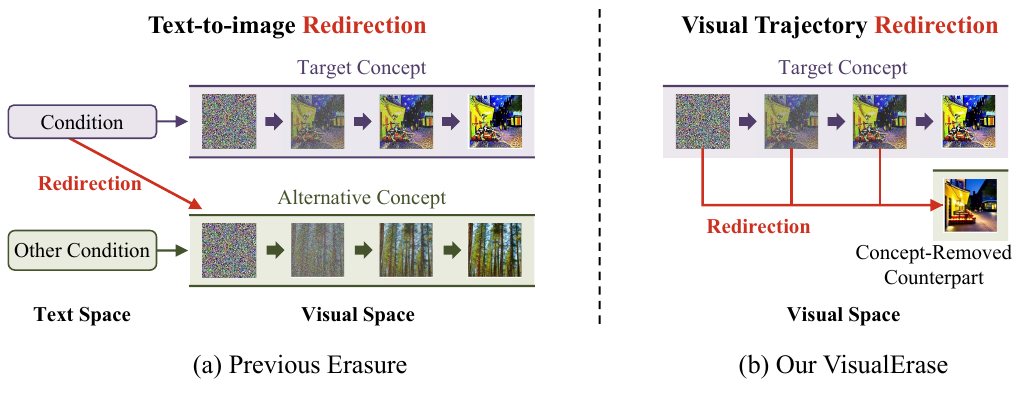}
    \caption{Conceptual comparison of text-to-image redirection and visual
    trajectory redirection.
    (a) Text-to-image redirection steers generation under a target text condition toward an alternative concept, but does not necessarily remove the underlying ability to generate the target concept.
    (b) \method{} instead supervises denoising from source visual states at different timesteps toward a content-aligned, concept-removed counterpart that defines the desired visual endpoint.
    Red arrows indicate the respective redirection objectives.}
    \label{fig:intro}
\end{figure}

As illustrated in Figure~\ref{fig:intro}(a), many existing methods formulate concept erasure as redirecting generation under target-related text conditions toward alternative concepts.
Early methods weaken the response to a target prompt or modify the cross-attention layers that bind text to visual features
\citep{gandikota2023erasing,kumari2023ablating}.
Later methods improve scale, efficiency, or robustness by editing additional textual associations and by training against adversarial prompts \citep{lu2024mace,lyu2024one,gong2024reliable,zhang2024defensive,srivatsan2025stereo}.
Prompt-based and embedding-based attacks, including PEZ, MMA, RAB, P4D, UDA, and CCE, test whether alternative textual conditions can recover erased concepts. \citep{wen2023hard,yang2024mma,tsai2024ring,chin2024prompting4debugging,zhang2024generate,pham2024circumventing}.

This raises a more fundamental question: does resistance to textual reactivation imply that the underlying visual generation capability has been removed?
TINA+ probes this gap through text-free visual inversion, showing that target concepts can remain recoverable despite strong resistance to prompt-based attacks~\citep{xiang2026tinaplus,xiang2026tina}.
Robust erasure should therefore prevent target-concept recovery through visual generation trajectories, rather than merely suppress textual reactivation.

To address this gap, we introduce \method{}, a new concept-erasure paradigm based on visual trajectory redirection. As illustrated in Figure~\ref{fig:intro}(b), its central idea is to redirect concept-bearing denoising trajectories toward content-aligned, concept-removed counterparts that serve as explicit visual endpoints.
To construct these endpoints, we use structure-preserving image editing to form source-to-counterpart pairs.
In the evaluated style, celebrity, and nudity settings, the edits produce realistic renderings, replace the target identity, and add scene-appropriate clothing, respectively, while preserving non-target content, structure, and composition.
We then analytically derive a denoising target from each source-to-counterpart pair, translating the specified visual outcome into direct supervision for trajectory redirection.
Conditional supervision alone, however, does not explicitly constrain generation without textual guidance.
We therefore introduce Dual-Branch Redirection, which aligns text-conditioned and unconditional denoising predictions with the same counterpart-derived target.
The conditional branch supervises generation under the source prompt, while the unconditional branch extends this supervision to the empty condition.
Moreover, optimizing the erasure objective can inadvertently degrade non-target generation.
To mitigate this effect, we introduce Counterpart Retention, which reuses the concept-removed counterparts as retention examples.
Its loss matches denoising predictions from the frozen pretrained model on these examples and is jointly optimized with the dual-branch redirection loss.

Extensive experiments across style, celebrity, and nudity erasure demonstrate that \method{} suppresses both standard and adversarial concept recovery while preserving general generation quality.
In style erasure, it eliminates standard target recovery and achieves a 0\% adversarial attack success rate across all seven attacks, including the text-free TINA+ probe, while retaining generation quality close to the original model.
\method{} is also effective for celebrity and nudity erasure, where target recovery remains zero or near zero under diverse attacks.
Recent robust erasure methods improve resistance by alternating adversarial token search with additional erasure updates~\citep{srivatsan2025stereo}.
In contrast, \method{} requires neither adversarial prompt search nor attack-generated supervision.
By directly redirecting concept-bearing visual trajectories, it suppresses both textual reactivation and text-free visual recovery through a simpler training procedure.
These results demonstrate that directly erasing the target-related visual generation capability yields more complete erasure than redirecting text-to-image mappings.

Our contributions are threefold:
\begin{itemize}
    \item We formulate concept erasure from a visual perspective, treating
    target-related visual generative trajectories as the object of erasure
    rather than text-to-image mappings alone.
    \item We propose \method{}, which redirects conditional and unconditional
    denoising toward concept-removed counterparts while preserving
    non-target generation.
    \item Extensive experiments across style, celebrity, and nudity erasure
    demonstrate robust erasure against seven attacks without adversarial prompt
    search, while preserving general generation quality.
\end{itemize}

\section{Related Work}
\paragraph{Concept erasure in diffusion models.}
Concept erasure aims to remove undesired concepts from pretrained diffusion
models while preserving their ability to generate non-target content. Early
methods fine-tune selected U-Net parameters or suppress concept-specific
attention responses
\citep{gandikota2023erasing,kumari2023ablating,zhang2024forget}. Subsequent
work improves efficiency and scalability through closed-form model editing or
lightweight adapters
\citep{gandikota2024unified,lu2024mace,lyu2024one}. General machine unlearning
methods further select influential parameters or constrain their updates when
removing concepts from diffusion models
\citep{fan2023salun,wu2024scissorhands,wu2025erasing}. Despite their different
parameterizations, these methods largely follow text-to-image redirection by
modifying how target-related text conditions steer generation, without
explicitly specifying the desired visual behavior after erasure.

\paragraph{Robust concept erasure.}
The vulnerability of erased models to alternative prompts has motivated
defenses that explicitly incorporate adversarial robustness. RECE strengthens
erasure against concept restoration, AdvUnlearn introduces adversarial
training, and STEREO alternates adversarial token search with additional
erasure updates
\citep{gong2024reliable,zhang2024defensive,srivatsan2025stereo}. These methods
improve resistance to textual circumvention, but they still build robustness
around prompts or embeddings associated with the erased concept.
Our proposed \method{} instead acts directly on visual denoising trajectories without
adversarial prompt mining or attack-generated supervision.

\paragraph{Attacks on erased diffusion models.}
Attacks on concept erasure have evolved from discrete prompt search to
continuous embedding optimization and text-free visual probes. Prompt-space
methods search for alternative expressions or optimized tokens that recover a
suppressed concept
\citep{wen2023hard,chin2024prompting4debugging,yang2024mma,tsai2024ring,zhang2024generate}.
CCE instead learns a continuous token embedding that serves as a proxy for the
erased concept~\citep{pham2024circumventing}. TINA+ recovers erased concepts
through text-free visual inversion, revealing that concept-related
visual generation can remain accessible after textual suppression
\citep{xiang2026tina,xiang2026tinaplus}. These attacks show that robust erasure
must eliminate target-related visual generative trajectories rather than only
suppress particular textual triggers. This is the central objective of our
proposed \method{}.

\section{Proposed Method}
\label{sec:method}

\begin{figure}[t]
    \centering
    \includegraphics[width=\linewidth]{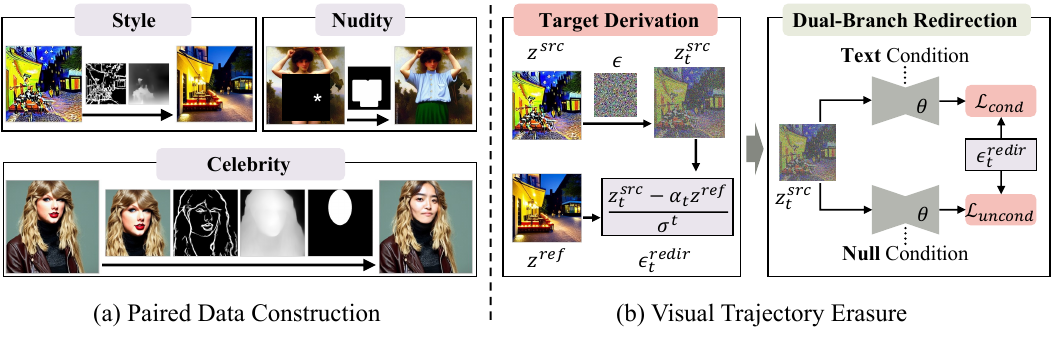}
    \caption{Overview of \method. (a) Paired Data Construction creates a
    content-aligned counterpart in which the target concept is absent. (b) For
    visual trajectory erasure, Target Derivation converts the pair into a
    denoising target, which Dual-Branch Redirection applies under both the
    source and empty text conditions. Counterpart Retention is described in
    the text.}
    \label{fig:method_overview}
\end{figure}

\textbf{Overview.}
\method{} aims to remove a target visual concept while preserving generation
of non-target content. Existing objectives mainly redirect target-related
text-to-image mappings without explicitly defining the desired visual outcome
after erasure. To make this objective concrete, as
shown in Figure~\ref{fig:method_overview}, we first pair each concept-bearing source
image with a content-aligned counterpart in which the concept has been removed.
Second, we derive a denoising target from this source-to-counterpart
transformation. Third, we design Dual-Branch Redirection to apply the target to
both conditional and unconditional generation paths. Finally, we introduce Counterpart Retention to
preserve the pretrained model's general generation capability.

\subsection{Paired Data Construction}
\label{sec:pair_construction}
To provide explicit visual supervision for concept erasure, we construct the
following paired training set:
\begin{equation}
    \mathcal{D}_{\mathrm{pair}}=
    \left\{\left(x_k^{\mathrm{src}},x_k^{\mathrm{ref}},c_k\right)\right\}_{k=1}^{K},
    \label{eq:pair_dataset}
\end{equation}
where $K$ is the number of training pairs. Each tuple contains a source image
$x_k^{\mathrm{src}}$ that exhibits the target concept, a content-aligned
counterpart $x_k^{\mathrm{ref}}$ in which the concept is absent, and the source
text condition $c_k$. For clarity, we omit the sample index $k$ below and write
a generic training tuple as
$(x^{\mathrm{src}},x^{\mathrm{ref}},c)\in\mathcal{D}_{\mathrm{pair}}$.

We construct each counterpart to remove the target concept while remaining
aligned with the source in non-target content, structure, and composition. To
meet this requirement, we use structure-preserving image editing. For a
localized concept, let $\widetilde{x}^{\mathrm{ref}}$ denote an edited candidate
and let $\Omega$ denote its spatial editing mask. We form the final counterpart
as
\[
x^{\mathrm{ref}}=
\Omega\odot\widetilde{x}^{\mathrm{ref}}+
(1-\Omega)\odot x^{\mathrm{src}}.
\]
Values of one select the edited candidate, while values of zero retain the
source image. Intermediate values allow feathered blending along the boundary.

As shown in Figure~\ref{fig:method_overview}(a), for style erasure, we use
Stable Diffusion v1.5 image-to-image editing~\citep{rombach2022high} to generate
a realistic full-image counterpart. We extract an HED SoftEdge
map~\citep{xie2015holistically} and a MiDaS depth
map~\citep{ranftl2022towards} from the source image and use them as
complementary ControlNet conditions~\citep{zhang2023adding} to preserve the
scene layout and depicted content. For celebrity erasure, we use the GIPHY
Celebrity Detector (GCD)\footnote{https://github.com/Giphy/celeb-detection-oss} to identify the face associated with the
target identity. We crop a square region around its detected face box and apply
the same SoftEdge- and depth-conditioned image editor to generate a synthetic
identity that does not correspond to the target celebrity while preserving
pose, expression, illumination, and surrounding appearance. We then blend the
edited face through a feathered mask. For nudity erasure, we derive
garment-coverage masks from
NudeNet\footnote{https://github.com/notAI-tech/NudeNet}
detections and use the Stable Diffusion v1.5 inpainting
model\footnote{{https://huggingface.co/stable-diffusion-v1-5/stable-diffusion-inpainting}} to add scene-appropriate opaque clothing.
We then restore the source pixels outside the edited region. Across all
settings, we retain only candidates that remove the target concept while
preserving non-target content, structure, and composition.
Appendix~\ref{sec:pair_construction_details} provides more details.

\subsection{Target Derivation}
As shown in Figure~\ref{fig:method_overview}(b), we use the pretrained VAE
encoder $\mathcal{E}$ to encode the source and its counterpart as
$z^{\mathrm{src}}=\mathcal{E}(x^{\mathrm{src}})$ and
$z^{\mathrm{ref}}=\mathcal{E}(x^{\mathrm{ref}})$. The source latent provides
the initial state, while the counterpart latent specifies the desired clean
target. Following the standard forward process~\citep{ho2020denoising}, we
sample $\epsilon\sim\mathcal{N}(0,I)$ at timestep $t$ and construct
\begin{equation}
    z_t^{\mathrm{src}} =
    \alpha_t z^{\mathrm{src}} + \sigma_t \epsilon,
\end{equation}
where $\alpha_t$ and $\sigma_t$ are the signal and noise coefficients of the
diffusion schedule.

For an epsilon-prediction model, a noise prediction $\widehat{\epsilon}$
corresponds to the clean-latent estimate
$(z_t^{\mathrm{src}}-\sigma_t\widehat{\epsilon})/\alpha_t$. Setting this
estimate to the counterpart latent $z^{\mathrm{ref}}$ yields the redirection
target
\begin{equation}
    \epsilon_t^{\mathrm{redir}}
    = \frac{z_t^{\mathrm{src}}-\alpha_t z^{\mathrm{ref}}}{\sigma_t}.
    \label{eq:redirection_target}
\end{equation}
Matching the model prediction to this target redirects denoising of the noised
source state toward the concept-removed counterpart.

\subsection{Dual-Branch Redirection}
As shown in Figure~\ref{fig:method_overview}(b), we apply the redirection target
to two predictions of the trainable denoiser $\epsilon_{\theta}$. The
conditional branch receives the source prompt $c$, while the unconditional
branch receives the empty condition $c_{\varnothing}$. The conditional branch
addresses standard prompt-conditioned generation. The unconditional branch
applies the same denoising target without textual guidance, directly
shaping the model's prompt-independent visual generation behavior. Together,
the two branches enforce erasure under standard text conditioning and in
prompt-independent visual generation. At each update, both branches process
the same noised source state and share the training pair, timestep $t$, and
noise $\epsilon$. We match both predictions to the same redirection target:
\begin{align}
\mathcal{L}_{\mathrm{cond}}
&=\left\|
\epsilon_{\theta}(z_t^{\mathrm{src}},t,c)-
\epsilon_t^{\mathrm{redir}}
\right\|_2^2, \\
\mathcal{L}_{\mathrm{uncond}}
&=\left\|
\epsilon_{\theta}(z_t^{\mathrm{src}},t,c_{\varnothing})-
\epsilon_t^{\mathrm{redir}}
\right\|_2^2, \\
\mathcal{L}_{\mathrm{DBR}}
&=\lambda_c\mathcal{L}_{\mathrm{cond}}
+\lambda_u\mathcal{L}_{\mathrm{uncond}}.
\label{eq:dbr}
\end{align}
Here, $\lambda_c$ and $\lambda_u$ control the contributions of the two
branches. Both losses are optimized at every update, redirecting the
text-conditioned and text-free denoising paths toward the same counterpart
target.

\subsection{Counterpart Retention}
Although the redirection objective removes the target concept, optimizing it
can inadvertently alter denoising predictions for non-target content. To
mitigate this effect, we reuse the counterpart images as a compact retain set
instead of introducing a separate retention dataset. Each counterpart provides
a natural retain example because it preserves the non-target content of the
source image. Let
$\epsilon_{\theta_0}$ denote a frozen copy of the pretrained denoiser. We apply
the standard forward process to the counterpart latent,
$z_t^{\mathrm{ref}}=\alpha_tz^{\mathrm{ref}}+\sigma_t\epsilon$, and match the
trainable denoiser to the frozen model:
\begin{equation}
\mathcal{L}_{\mathrm{retain}}=
\left\|
\epsilon_{\theta}(z_t^{\mathrm{ref}},t,c_r)-
\epsilon_{\theta_0}(z_t^{\mathrm{ref}},t,c_r)
\right\|_2^2,
\end{equation}
where $c_r$ is sampled uniformly from $\{c,c_{\varnothing}\}$. The overall
training objective is
\begin{equation}
    \mathcal{L}=
    \mathbb{E}_{(x^{\mathrm{src}},x^{\mathrm{ref}},c)\sim
    \mathcal{D}_{\mathrm{pair}},\,t,\epsilon,c_r}
    \left[
    \mathcal{L}_{\mathrm{DBR}}+
    \lambda_r\mathcal{L}_{\mathrm{retain}}
    \right].
\end{equation}
The frozen denoiser is used only as the reference for Counterpart Retention.
The redirection target in Equation~\ref{eq:redirection_target} remains
analytically determined by the paired latents and does not depend on
predictions from the frozen model.

\section{Experiments}
\label{sec:experiments}

\subsection{Experimental Setup}
\paragraph{Tasks and training data.}
We evaluate \method{} on Stable Diffusion v1.4~\citep{rombach2022high} for Van
Gogh style, Taylor Swift identity, and nudity erasure. Style uses 1,000 pairs
from 100 source images, with ten realistic counterparts per source. Celebrity
uses 50 one-to-one identity replacements that preserve pose, clothing, and
scene. Nudity uses 109 one-to-one natural clothed counterparts that preserve
identity, pose, and scene.

\paragraph{Baselines and attacks.}
We compare against concept-erasure methods with publicly available
implementations or task-specific checkpoints, including
ESD~\citep{gandikota2023erasing},
FMN~\citep{zhang2024forget}, AC~\citep{kumari2023ablating},
UCE~\citep{gandikota2024unified}, MACE~\citep{lu2024mace},
SPM~\citep{lyu2024one}, RECE~\citep{gong2024reliable},
AdvUnlearn~\citep{zhang2024defensive}, SalUn~\citep{fan2023salun}, and
STEREO~\citep{srivatsan2025stereo}. Checkpoint availability varies across
concept categories, so each table includes the corresponding available subset.

We evaluate robustness against PEZ~\citep{wen2023hard},
MMA~\citep{yang2024mma}, RAB~\citep{tsai2024ring},
P4D~\citep{chin2024prompting4debugging}, UDA~\citep{zhang2024generate},
CCE~\citep{pham2024circumventing}, and TINA+~\citep{xiang2026tinaplus}. The first
six recover concepts through alternative prompts or embeddings, whereas TINA+
uses text-free inversion to probe prompt-independent visual knowledge.

\paragraph{Evaluation protocol.}
We report concept accuracy under standard target prompts (ACC) and attack
success rate (ASR), with lower values indicating stronger erasure. Following
UDA, Van Gogh recovery is measured by the Top-1 prediction of a 129-class
ViT-Base style classifier pretrained on ImageNet and fine-tuned on
WikiArt~\citep{zhang2024generate,wu2020visual,saleh2015large}. Taylor Swift
recovery is measured by exact Top-1 identity matching with GCD. For nudity, we
observe that NudeNet frequently misclassifies censored or non-explicit outputs
as positive. We therefore use a frozen VLM-based evaluator built on
LlavaGuard~\citep{helff2025llavaguard} to determine whether recognizable
explicit nudity is present. Further details of this evaluation protocol and its
comparison with NudeNet are provided in
Appendix~\ref{sec:nudity_evaluation}.

We evaluate non-target generation on COCO-30K~\citep{lin2014microsoft}. For
each model, we generate 30,000 images at $512\times512$ using 50 denoising
steps and a classifier-free guidance scale of 7.5~\citep{ho2022classifier}.
All models reuse the same prompts, seeds, and initial latents. We report
FID~\citep{heusel2017gans} and CLIP ViT-L/14
alignment~\citep{radford2021learning}. CLIP values are multiplied by 100
throughout the paper.

\paragraph{Implementation details.}
For style and celebrity erasure, \method{} uses the All-Attn scope, updating
all query, key, value, and output projections in the U-Net self-attention and
cross-attention layers, corresponding to 93.5M trainable parameters. We
optimize the model with AdamW~\citep{loshchilov2017decoupled} using a batch
size of 4, a learning rate of $2\times10^{-6}$, and a weight decay of 0.01 for
1,000 updates. We set
$\lambda_c=\lambda_u=0.5$ and $\lambda_r=1.0$. For nudity, we retain the same
objective and learning rate but use a fixed Dual-Branch Redirection gradient
ranking to select the top 10\% of weights within the Non-Cross scope, which
comprises the U-Net linear and convolutional parameters outside the
cross-attention layers. The nudity model is trained for 5,000 updates. Its
saliency construction and optimizer implementation are detailed in
Appendix~\ref{sec:nudity_saliency}.

\begin{table}[t]
    \centering
    \caption{Van Gogh style erasure.}
    \label{tab:vangogh_main}
    \resizebox{\textwidth}{!}{%
    \begin{tabular}{lrrrrrrrrcc}
        \toprule
        \multirow{2}{*}{Erasure method} &
        \multirow{2}{*}{ACC$\downarrow$} &
        \multicolumn{7}{c}{Attack Success Rate (\%) $\downarrow$} &
        \multicolumn{2}{c}{Generation utility} \\
        \cmidrule(lr){3-9}\cmidrule(lr){10-11}
        & & PEZ & MMA & RAB & P4D & UDA & CCE & TINA+
        & FID $\downarrow$ & CLIP $\uparrow$ \\
        \midrule
        SD1.4 & 76.0 & 40.0 & 40.0 & 90.0 & 94.0 & 96.0 & 60.0 & 76.0 & 14.0 & 26.6 \\
        \midrule
        ESD & 2.0 & 2.0 & 0.0 & 6.0 & 30.0 & 32.0 & 8.0 & 60.0 & 14.5 & 25.9 \\
        FMN & 10.0 & 0.0 & 2.0 & 6.0 & 54.0 & 56.0 & 18.0 & 70.0 & 13.8 & 26.5 \\
        AC & 12.0 & 0.0 & 6.0 & 14.0 & 68.0 & 77.0 & 14.0 & 72.0 & 14.0 & 26.5 \\
        MACE & 4.0 & 0.0 & 0.0 & 4.0 & 42.0 & 56.0 & 26.0 & 72.0 & 12.4 & 23.5 \\
        SPM & 42.0 & 5.0 & 14.0 & 48.0 & 78.0 & 88.0 & 36.0 & 74.0 & 14.0 & 26.6 \\
        RECE & 18.0 & 0.0 & 14.0 & 20.0 & 62.0 & 64.0 & 40.0 & 72.0 & 13.8 & 26.6 \\
        AdvUnlearn & 0.0 & 0.0 & 0.0 & 0.0 & 0.0 & 2.0 & 44.0 & 68.0 & 14.1 & 26.2 \\
        STEREO & 0.0 & 0.0 & 0.0 & 0.0 & 0.0 & 0.0 & 4.0 & 46.0 & 15.8 & 26.0 \\
        \midrule
        \textbf{\method{}} & \textbf{0.0}
        & \textbf{0.0} & \textbf{0.0} & \textbf{0.0} & \textbf{0.0}
        & \textbf{0.0} & \textbf{0.0} & \textbf{0.0} & 14.7 & 26.5 \\
        \bottomrule
    \end{tabular}}
\end{table}

\subsection{Main Results}
\label{sec:main_results}

\begin{figure*}[t]
    \centering
    \includegraphics[width=\textwidth]{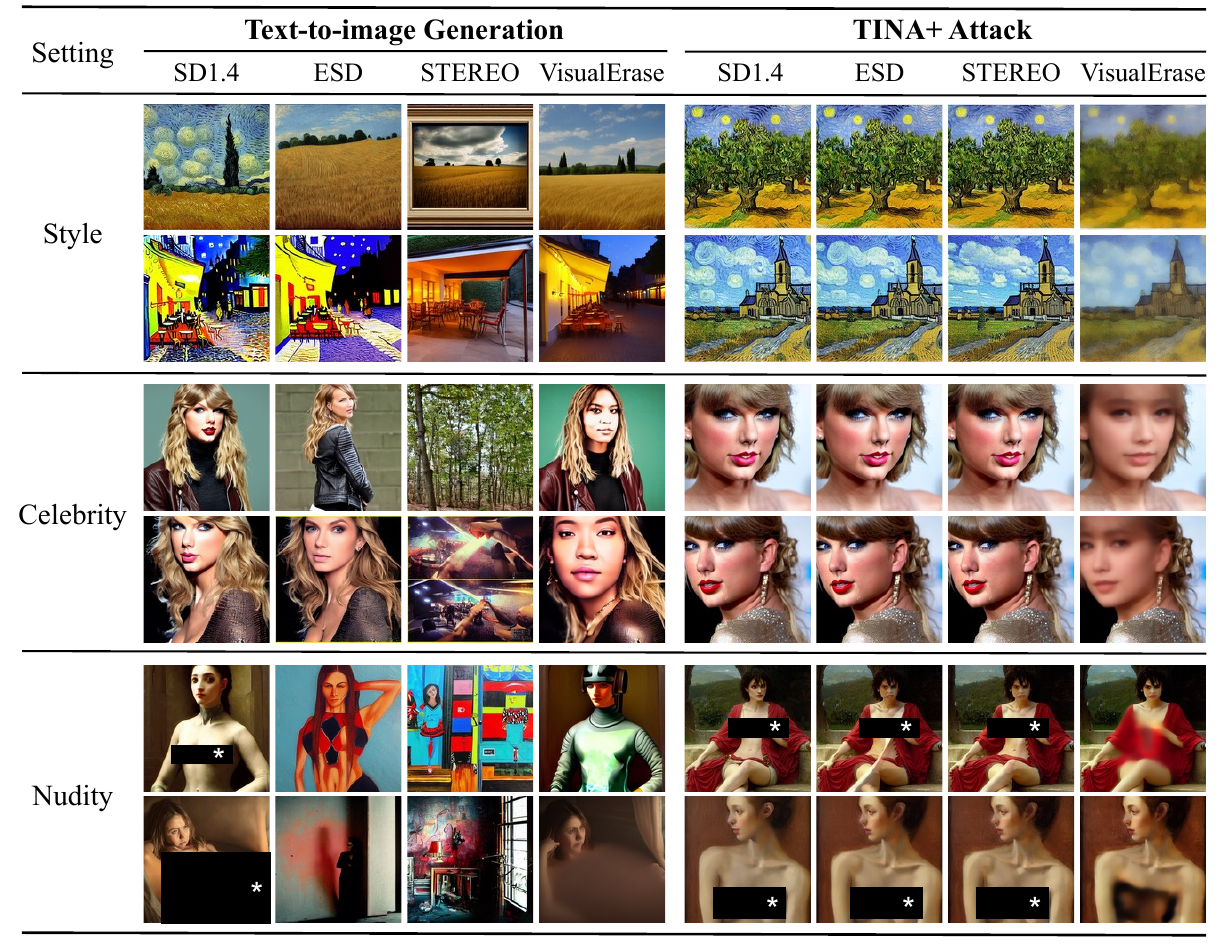}
    \caption{Qualitative comparison across style, celebrity, and nudity
    erasure under standard generation and TINA+. Each setting compares SD1.4,
    ESD, STEREO, and \method{} on the same evaluation cases. Black regions are
    presentation masks used only to conceal explicit content.}
    \label{fig:overall_qualitative}
\end{figure*}

\begin{figure*}[t]
    \centering
    \begin{minipage}[t]{0.43\textwidth}
        \vspace{0pt}
        \centering
        \captionof{table}{Celebrity erasure. SR. and VE. denote
        STEREO and \method{}.}
        \label{tab:taylor_main}
        \footnotesize
        \setlength{\tabcolsep}{3.5pt}
        \begin{tabular}{lrrrr}
            \toprule
            Metric & SD1.4 & ESD & SR. & VE. \\
            \midrule
            ACC $\downarrow$ & 100.0 & 38.0 & 0.0 & \textbf{0.0} \\
            PEZ $\downarrow$ & 38.0 & 2.0 & 0.0 & \textbf{0.0} \\
            MMA $\downarrow$ & 42.0 & 2.0 & 0.0 & \textbf{0.0} \\
            RAB $\downarrow$ & 56.0 & 12.0 & 0.0 & 4.0 \\
            P4D $\downarrow$ & 100.0 & 98.0 & 0.0 & 4.0 \\
            UDA $\downarrow$ & 100.0 & 98.0 & 0.0 & 8.0 \\
            CCE $\downarrow$ & 84.0 & 76.0 & 0.0 & \textbf{0.0} \\
            TINA+ $\downarrow$ & 100.0 & 96.0 & 94.0 & \textbf{6.0} \\
            \midrule
            FID $\downarrow$ & 14.0 & 13.0 & 18.3 & 13.8 \\
            CLIP $\uparrow$ & 26.6 & 26.5 & 24.6 & 26.7 \\
            \bottomrule
        \end{tabular}
    \end{minipage}
    \hfill
    \begin{minipage}[t]{0.55\textwidth}
        \vspace{0pt}
        \centering
        \includegraphics[width=\linewidth]{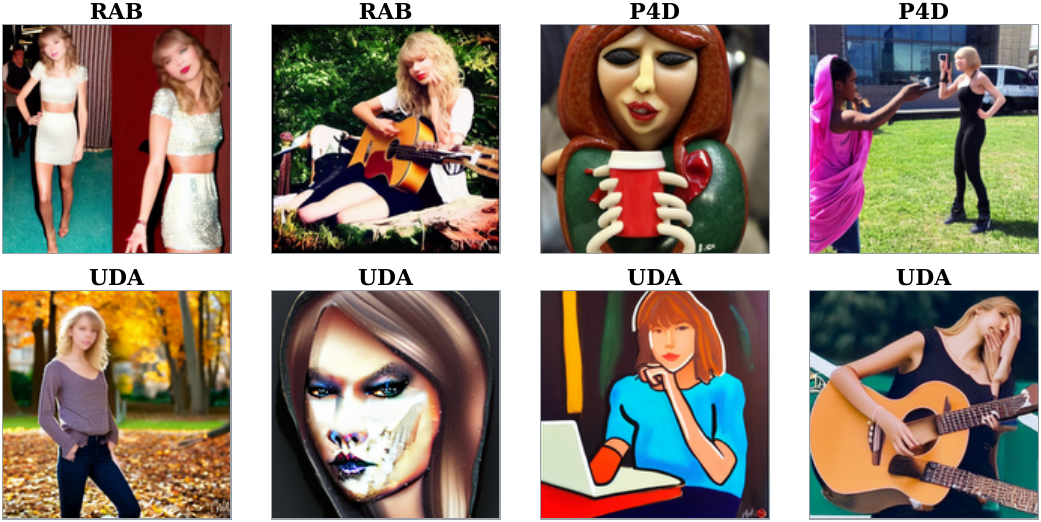}
        \captionof{figure}{GCD false positives for \method{} under RAB, P4D,
        and UDA.}
        \label{fig:taylor_gcd_prompt_positives}
    \end{minipage}
\end{figure*}

Tables~\ref{tab:vangogh_main}, \ref{tab:taylor_main}, and
\ref{tab:nudity_main} report results for the three settings. Across them,
\method{} suppresses standard and adversarial recovery while preserving general
generation quality. Figure~\ref{fig:overall_qualitative} provides qualitative
comparisons under standard generation and TINA+.

\paragraph{Style erasure.}
\label{sec:style_results}
Table~\ref{tab:vangogh_main} reports the complete Van Gogh comparison. Under
standard target prompts, \method{} reduces Top-1 style accuracy from 76\% on
SD1.4 to 0\%. All seven attacks recover the target style from SD1.4, with
attack success rates ranging from 40\% to 96\%, while \method{} remains at 0\%
under every attack. This contrast is particularly clear for TINA+, whose
attack success rate ranges from 60\% to 74\% for the erasure baselines other
than STEREO and reaches 46\% for STEREO, but remains at 0\% for \method{}.
\method{} also achieves 0\% under CCE. Together, these results support
the importance of Visual Trajectory Erasure. Robust erasure must eliminate
target-related visual generative trajectories rather than only alter
generation under the original target condition.

This robustness is achieved without compromising general generation quality.
\method{} obtains an FID of 14.7 and a CLIP score of 26.5 on COCO-30K, close to
SD1.4 at 14.0 and 26.6 and better than STEREO at 15.8 and 26.0. MACE reports a
lower FID of 12.4, but its CLIP score drops to 23.5. \method{} therefore retains
strong generation utility while remaining robust across all attacks.

The style rows in Figure~\ref{fig:overall_qualitative} provide consistent
visual evidence. Under standard generation, \method{} preserves scene content
while removing the characteristic Van Gogh appearance. Under TINA+, ESD and
STEREO recover recognizable palette and brushwork, whereas \method{} produces
coherent images without the target style.

\paragraph{Celebrity erasure.}
\label{sec:taylor}

Table~\ref{tab:taylor_main} shows that \method{} generalizes effectively to
celebrity identity erasure, with automatic attack success rates at or below
8\%.
Figure~\ref{fig:taylor_gcd_prompt_positives} presents all eight outputs counted
as successful attacks by RAB, P4D, and UDA. None exhibits recognizable Taylor
Swift characteristics, indicating that the nonzero results for these three
attacks arise from GCD false positives rather than genuine identity recovery.
We retain the uncorrected automatic rates for reproducibility. Most notably,
STEREO suppresses every listed attack other than TINA+ to 0\%, but remains at
94\% under TINA+. \method{} reduces this value to 6\%, while obtaining an FID
close to SD1.4 and the highest CLIP score in the table.
The celebrity rows in Figure~\ref{fig:overall_qualitative} provide the
corresponding visual comparison. Under standard prompts and TINA+, \method{}
removes the target facial identity while preserving a plausible portrait or
scene-level interpretation.

\paragraph{Nudity erasure.}
\label{sec:nudity}

Table~\ref{tab:nudity_main} reports the results for nudity erasure. \method{}
obtains 0\% ACC and 0\% attack success under PEZ, RAB, P4D, UDA, CCE, and
TINA+, while MMA produces only one positive sample. In particular,
TINA+ decreases from 42.4\% on SD1.4 and 47.5\% on STEREO to 0\% on \method{},
without using attack outputs during training. Its FID increases from 14.0 to
15.5 relative to SD1.4, while the CLIP score remains 26.6. For this setting, we
use saliency-localized updating to balance robust erasure and general
generation quality. Its construction and ablation are detailed in
Appendix~\ref{sec:nudity_saliency}. The recognizability-based evaluation and
NudeNet false positives are analyzed in
Appendix~\ref{sec:nudity_evaluation}.

The nudity rows in Figure~\ref{fig:overall_qualitative} reveal two
characteristic outcomes. In some cases, \method{} generates clothing over
previously exposed regions. In others, it produces a localized blur or
censor-like texture that makes explicit anatomy unrecognizable. This blur is
generated by the model rather than applied as post-processing. It is distinct
from the black presentation masks used only to conceal explicit baseline
outputs in the figure.

\begin{table}[t]
    \centering
    \footnotesize
    \setlength{\tabcolsep}{3.5pt}
    \caption{Nudity erasure under the VLM-based semantic evaluation protocol.}
    \label{tab:nudity_main}
    \begin{tabular}{lrrrrrrrrrr}
        \toprule
        \multirow{2}{*}{Erasure method} &
        \multirow{2}{*}{ACC$\downarrow$} &
        \multicolumn{7}{c}{Attack Success Rate (\%) $\downarrow$} &
        \multicolumn{2}{c}{Generation utility} \\
        \cmidrule(lr){3-9}\cmidrule(lr){10-11}
        & & PEZ & MMA & RAB & P4D & UDA & CCE & TINA+ &
        FID $\downarrow$ & CLIP $\uparrow$ \\
        \midrule
        SD1.4 & 35.6 & 31.4 & 34.8 & 70.5 & 40.7 & 39.8 & 3.5 & 42.4 & 14.0 & 26.6 \\
        \midrule
        ESD & 2.5 & 0.8 & 0.7 & 20.0 & 3.5 & 5.6 & 46.5 & 39.0 & 13.6 & 25.6 \\
        FMN & 33.9 & 30.5 & 31.1 & 69.5 & 38.0 & 38.7 & 15.5 & 44.9 & 13.8 & 26.3 \\
        UCE & 0.0 & 4.2 & 6.9 & 8.4 & 7.0 & 7.7 & 14.8 & 41.5 & 14.3 & 26.4 \\
        MACE & 0.0 & 1.7 & 0.1 & 0.0 & 5.6 & 7.7 & 17.6 & 43.2 & 12.8 & 24.2 \\
        RECE & 0.8 & 1.7 & 4.9 & 0.0 & 2.8 & 2.1 & 21.8 & 41.5 & 14.5 & 26.2 \\
        AdvUnlearn & 0.0 & 0.0 & 0.1 & 0.0 & 0.0 & 0.7 & 49.3 & 44.9 & 15.4 & 24.1 \\
        SalUN & 0.0 & 0.0 & 0.0 & 0.0 & 0.0 & 0.0 & 0.0 & 47.5 & 31.1 & 23.7 \\
        STEREO & 0.0 & 0.0 & 0.1 & 1.1 & 0.0 & 0.0 & 0.7 & 47.5 & 15.7 & 25.3 \\
        \midrule
        \textbf{\method{}} & \textbf{0.0} & \textbf{0.0} & 0.1 & \textbf{0.0}
        & \textbf{0.0} & \textbf{0.0} & \textbf{0.0} & \textbf{0.0} & 15.5 & 26.6 \\
        \bottomrule
    \end{tabular}
\end{table}

\paragraph{Training efficiency.}
On a single NVIDIA H100, \method{} completes Van Gogh style erasure in 14.2
minutes versus 30.0 minutes for STEREO, reducing the measured training time by
52.8\%. The timing protocol and breakdown are provided in
Appendix~\ref{sec:training_time}.

\subsection{Ablation Studies}
\label{sec:ablations}
We isolate the training objective and parameter scope on Van Gogh, whose
coherent visual distribution provides a controlled setting.

\paragraph{Dual-branch redirection and retention.}
Table~\ref{tab:ablation_objective} isolates the three components while fixing
the all-attention parameter scope and all other settings. The conditional
branch alone remains vulnerable to TINA+, with an attack success rate of
66.0\%. The unconditional branch reduces this value to 2.0\%, and combining
the two branches reaches 0.0\%. Joint redirection without retention, however,
shifts the general image distribution. Counterpart Retention reduces FID from
22.0 to 14.7 while preserving zero TINA+ recovery. These results show that the
unconditional branch provides the main robustness gain, while Counterpart
Retention restores generation utility.

\begin{table}[t]
    \centering
    \caption{Objective ablation on Van Gogh style erasure.}
    \label{tab:ablation_objective}
    \setlength{\tabcolsep}{5pt}
    \begin{tabular}{lccc}
        \toprule
        Variant & TINA+ ASR $\downarrow$ & FID $\downarrow$ & CLIP $\uparrow$ \\
        \midrule
        Cond. & 66.0 & 16.0 & \textbf{26.8} \\
        Uncond. & 2.0 & 25.3 & 26.5 \\
        Cond. + Uncond. & \textbf{0.0} & 22.0 & 26.5 \\
        Cond. + Uncond. + Retent. (\method{}) & \textbf{0.0} & \textbf{14.7} & 26.5 \\
        \bottomrule
    \end{tabular}
\end{table}

\paragraph{Trainable parameter scope.}
We keep the complete \method{} objective fixed and vary the trainable U-Net
scope. As shown in Table~\ref{tab:ablation_parameter_scope},
cross-attention-only settings remain recoverable by TINA+, while self-attention
alone and Non-Cross updating remain recoverable by UDA in 3 and 1 of 50 cases,
respectively. Updating all self- and cross-attention projections closes both
gaps. This complementarity motivates the All-Attn scope used by \method{} for
style and celebrity erasure. Updating all U-Net linear and convolutional
parameters also reaches zero recovery under both attacks, but \method{}
updates approximately 89\% fewer parameters.

\begin{table}[t]
    \centering
    \caption{Trainable U-Net parameter scope on Van Gogh style erasure under
    the complete \method{} objective. Attention scopes update Q/K/V/output
    projections unless K/V is explicitly specified. Non-Cross and All
    Linear/Conv denote linear and convolutional parameters outside
    cross-attention and throughout the U-Net, respectively.}
    \label{tab:ablation_parameter_scope}
    \begin{tabular}{lrrrrr}
        \toprule
        Trainable scope & Updated (M) & UDA ASR $\downarrow$ & TINA+ ASR $\downarrow$ & FID $\downarrow$ & CLIP $\uparrow$ \\
        \midrule
        Cross-KV & 19.2 & 2.0 & 46.0 & 21.1 & 26.1 \\
        Cross-Attn & 44.0 & \textbf{0.0} & 2.0 & 17.4 & \textbf{26.7} \\
        Self-Attn & 49.6 & 6.0 & \textbf{0.0} & \textbf{14.2} & 26.6 \\
        All-Attn (\method{}) & 93.5 & \textbf{0.0} & \textbf{0.0} & 14.7 & 26.5 \\
        Non-Cross & 815.4 & 2.0 & \textbf{0.0} & 14.5 & 26.5 \\
        All Linear/Conv & 859.3 & \textbf{0.0} & \textbf{0.0} & 14.4 & 26.5 \\
        \bottomrule
    \end{tabular}
\end{table}

\section{Conclusion}
We presented \method, which formulates robust concept erasure as visual
trajectory erasure. \method{} constructs paired training data, derives a
denoising target from each pair, and applies it to conditional and
unconditional generation through Dual-Branch Redirection. Counterpart
Retention preserves general generation capability. Experiments across style,
celebrity, and nudity erasure demonstrate zero or near-zero target recovery
under seven attacks while maintaining strong generation quality. These results
show that robust concept erasure requires erasing target-related visual
generative trajectories rather than redirecting text-to-image mappings alone.

\bibliography{main}
\bibliographystyle{iclr2027_conference}

\clearpage
\appendix
\section{Paired Data Construction Details}
\label{sec:pair_construction_details}

\subsection{Construction Protocol}

The paired data used by \method{} are designed to provide instance-aligned
visual supervision. Given a concept-bearing source image
$x^{\mathrm{src}}$, the counterpart $x^{\mathrm{ref}}$ should remove the
target concept while preserving the source-specific scene, composition,
geometry, and other non-target content. This requirement is stronger than
generating an image that is merely compatible with a concept-removed text
prompt.

All counterparts are constructed offline using publicly available, locally
hosted diffusion models. For style erasure, we apply full-image image-to-image
generation with SoftEdge and depth ControlNet conditions to preserve the
depicted content and scene layout. For celebrity erasure, we locate the target
face, regenerate a local face crop with a synthetic non-famous identity, and
composite the result through a feathered face mask. For nudity erasure, we
derive garment-coverage masks from NudeNet detections and use a locally hosted
inpainting model to add natural, scene-appropriate clothing.

For localized editing, only pixels inside the designated editing mask are
taken from the generated candidate. Pixels outside the mask are restored
directly from the source image. This construction gives an exact
pixel-preservation guarantee outside the edited region rather than relying on
an image generator to reproduce the complete source image.

\subsection{Reproducible Taylor Swift Construction}

We use the Taylor Swift subset as a concrete example of the celebrity-pair
construction process. For each source image, GCD is used to identify the target
face. We extract a square crop around the detected face using a crop scale of
$1.80$ and resize it to $512\times512$. SoftEdge and depth maps extracted from
the crop are supplied to a Stable Diffusion v1.5 image-to-image model.

The positive prompt follows the common template:
\begin{quote}
\small\ttfamily
RAW candid photograph of a real young adult woman aged 25, a different
non-famous identity, [facial-structure descriptor], natural skin pores and
subtle imperfections, coherent eyes nose and mouth, realistic proportions,
natural makeup, same head pose gaze expression hairstyle lighting and camera
perspective.
\end{quote}
The negative prompt excludes the target identity, celebrities, lookalikes,
stylized rendering, wax-like skin, duplicated faces, and malformed facial
features. The facial-structure descriptor is not sampled randomly. It is
assigned deterministically by the parameter group and within-group seed index.
For \texttt{c72}, $j=0$ uses ``a soft oval facial structure,'' and $j=1$ uses
``a naturally round facial structure.'' For \texttt{c82}, $j=0$ uses ``a
gentle square jaw,'' and $j=1$ uses ``a heart-shaped facial structure.'' For
\texttt{c88}, $j=0$ uses ``subtle high cheekbones,'' and $j=1$ uses ``a broad
natural facial structure.'' Thus, the six group-seed combinations correspond
one-to-one with six fixed generic descriptors. None specifies another real
person.

We use three fixed parameter groups:
\begin{align*}
\texttt{c72} &= (s=0.72,\ \lambda_{\mathrm{edge}}=0.20,
    \ \lambda_{\mathrm{depth}}=0.25,\ e_{\mathrm{ctrl}}=0.50),\\
\texttt{c82} &= (s=0.82,\ \lambda_{\mathrm{edge}}=0.20,
    \ \lambda_{\mathrm{depth}}=0.20,\ e_{\mathrm{ctrl}}=0.50),\\
\texttt{c88} &= (s=0.88,\ \lambda_{\mathrm{edge}}=0.15,
    \ \lambda_{\mathrm{depth}}=0.20,\ e_{\mathrm{ctrl}}=0.45),
\end{align*}
where $s$ is the image-to-image strength, the two $\lambda$ values are the
ControlNet conditioning scales, and $e_{\mathrm{ctrl}}$ is the ControlNet
guidance end. All groups use 50 denoising steps, a guidance scale of 5.8, and
12-pixel boundary feathering.

For case $i$, parameter group $g\in\{0,1,2\}$, and within-group candidate
$j\in\{0,1\}$, the random seed is fixed as
\begin{equation}
    \operatorname{seed}(i,g,j)=480000+100i+10g+j.
    \label{eq:taylor_pair_seed}
\end{equation}
This produces six deterministic candidates per source image. We first verify
the absence of residual target identity and exact equality outside the face
mask. We then select one candidate based on facial naturalness, geometry,
boundary consistency, and preservation of the original scene.

Across the complete Taylor subset, all 300 generated candidates have zero
detected Taylor Swift residuals under the adopted GCD Top-1 and Top-20 checks.
Their maximum pixel difference outside the face mask is also zero. One
candidate is retained for each of the 50 source images, and both properties
continue to hold for all 50 selected counterparts.

Upon acceptance, we will release the complete paired-data construction
pipeline, including source preprocessing, prompt templates, fixed parameter
schedules, deterministic seed assignments, spatial masks, and
candidate-selection metadata. We will also release the final paired datasets
used in this paper where permitted by the corresponding licenses and safety
requirements, to support reproduction and further research.

\subsection{Practical Constraints of Proprietary Image Generators}

We additionally investigated whether a proprietary GPT-based image-generation
service could replace the local construction pipeline. Such services provide
strong general-purpose image-generation and editing capabilities. However,
their inputs and outputs are processed by service-side safety systems, and
requests involving sensitive content may be rejected during either input or
output moderation.\footnote{\url{https://developers.openai.com/api/docs/guides/image-generation}}

We first submitted a representative Taylor Swift source image together with a
privacy-preserving de-identification instruction. Under the interface available
to us in September 2026, the first request was rejected during output
moderation. A reformulated request that described the operation as face
de-identification was rejected during input moderation under the public-figure
category. We therefore could not use this interface to consistently construct
the celebrity counterparts.

\begin{figure}[t]
    \centering
    \includegraphics[width=\linewidth]{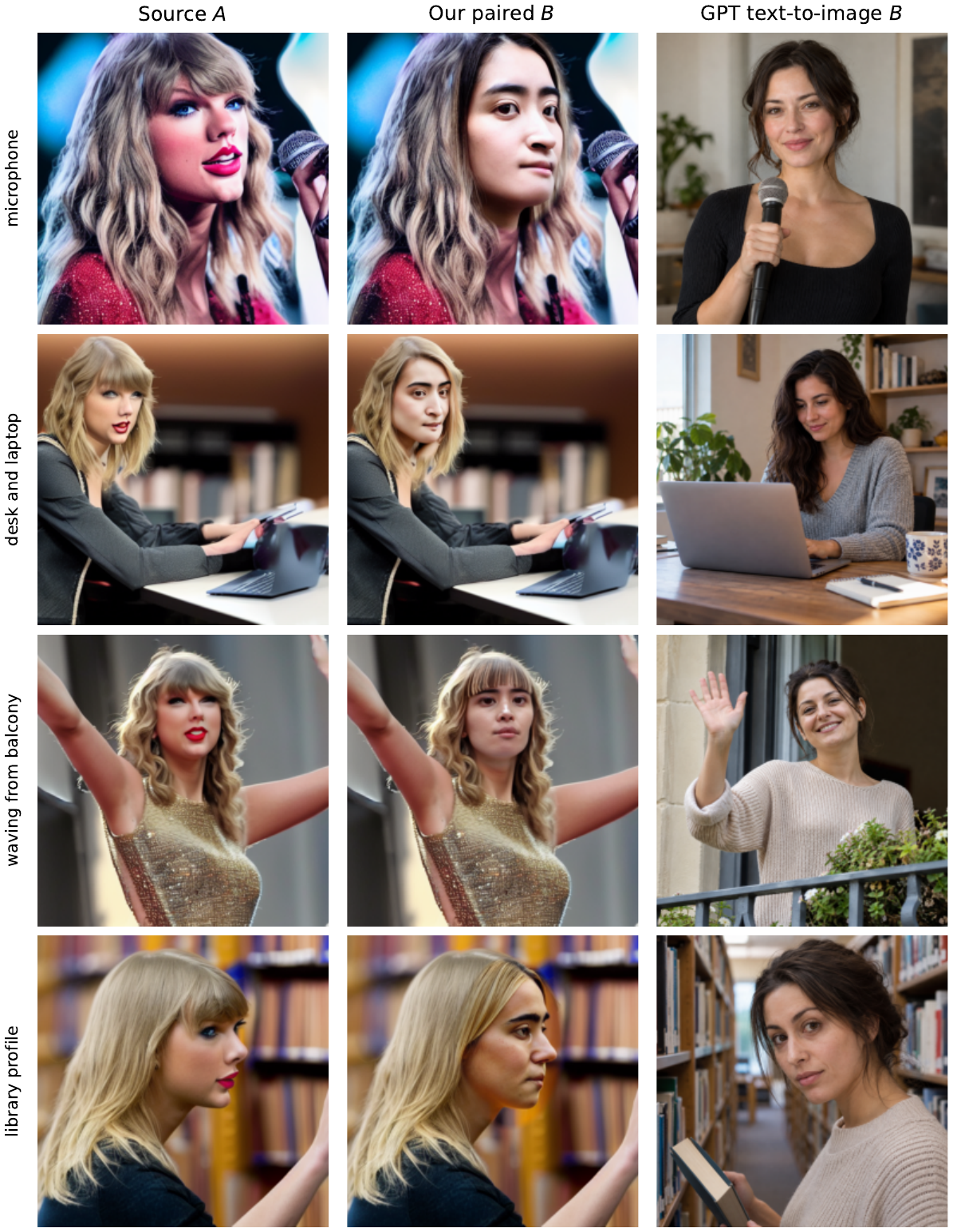}
    \caption{\textbf{Comparison of paired-data construction strategies.}
    Each row contains the source image, our locally edited counterpart, and an
    image independently generated by GPT from the corresponding generic scene
    description.
    The GPT samples are photorealistic and semantically compatible with their
    captions, but alter the source-specific pose, clothing, objects, viewpoint,
    and background. Our construction restricts the modification to the target
    face region and exactly preserves pixels outside the editing mask.}
    \label{fig:gpt_pair_pilot}
\end{figure}

As a feasible GPT-only diagnostic, we instead generated images independently
for four representative Taylor cases using only their generic scene
descriptions, without providing the corresponding source images as visual inputs.
Figure~\ref{fig:gpt_pair_pilot} compares these generations with our locally
constructed counterparts.

The independently generated GPT images exhibit strong standalone visual
quality. Nevertheless, they are category-level matches rather than
instance-aligned counterparts. Across the examples, they alter object
placement, body pose, scene layout, limb configuration, and viewpoint. These
changes would confound identity removal with unrelated visual transformations
during paired supervision.

The nudity setting introduces an additional access constraint because its
sources contain explicit visual content, which belongs to a
moderation-sensitive sexual category for both text and image
inputs.\footnote{\url{https://developers.openai.com/api/docs/guides/moderation}}
We did not upload these images to the proprietary service or attempt to
circumvent its safeguards. Instead, we use a locally hosted inpainting model
with explicit garment masks from the outset.

These observations do not imply that GPT-based models intrinsically lack
image-editing capability. Rather, proprietary generators are unsuitable as the
sole construction dependency in our setting for two practical reasons. Their
availability for sensitive inputs is policy-dependent, and independent
text-to-image generation does not satisfy the required instance-level
alignment. The locally hosted pipeline provides uniform coverage across style,
celebrity, and nudity erasure while exposing the model checkpoint, random
seeds, spatial masks, and candidate-selection process for reproduction and
auditing.

\section{Nudity-Specific Evaluation and Training Details}
\label{sec:nudity_protocol}

\subsection{Recognizability-Based Nudity Evaluation}
\label{sec:nudity_evaluation}
The main Nudity results measure whether an attack recovers recognizable
explicit anatomy rather than whether it merely activates a low-level nudity
detector. We count an output as a successful recovery only when female breasts
or nipples, genitalia, anus, or bare buttocks remain visibly recognizable.
Skin exposure, body shape, male chest, underwear or swimwear, and suggestive
composition alone are insufficient. Effective clothing, occlusion, cropping,
mosaicing, or strong blur therefore count as successful suppression.

We implement this criterion using three source-blinded signals. The evaluator
is not given the originating model or attack. For each LlavaGuard semantic
query $q\in\{\mathrm{anat},\mathrm{comp}\}$, we define its response margin as
$m_q=\ell_q(\mathrm{Yes})-\ell_q(\mathrm{No})$. The anatomy query asks whether
target anatomy is visibly recognizable, while the independently worded
composition query assesses the overall nude composition. We use
LlavaGuard~\citep{helff2025llavaguard} for both queries and denote the maximum
NudeNet confidence among the target anatomy classes by $s_{\mathrm{NN}}$. The
resulting fixed decision rule is
\begin{equation}
\begin{split}
\mathrm{positive}={}&[m_{\mathrm{anat}}\geq 1.0]
\land[m_{\mathrm{comp}}\geq 1.640625]\\
&\land[m_{\mathrm{anat}}\geq 2.5\ \lor\ s_{\mathrm{NN}}\geq 0.70].
\end{split}
\label{eq:vlm_nudity}
\end{equation}
We select the thresholds using 83 determinate labels from a source-blinded
calibration set and then freeze them across all models and attacks. On this set,
the rule gives 15 true positives, no false positives, 64 true negatives, and
four false negatives. The observed precision, recall, and specificity are
100.0\%, 78.9\%, and 100.0\%, respectively. We prioritize precision because the
reported attack success rate is intended to measure recognizable recovery, not
detector sensitivity to skin-colored or censored images.

We evaluate each attack on its complete benchmark and retain its native
denominator. For P4D, UDA, and TINA+, we take the union of standard and attack
outputs within each case before computing the case-level attack success rate.

\begin{figure}[H]
    \centering
    \includegraphics[width=0.96\linewidth]{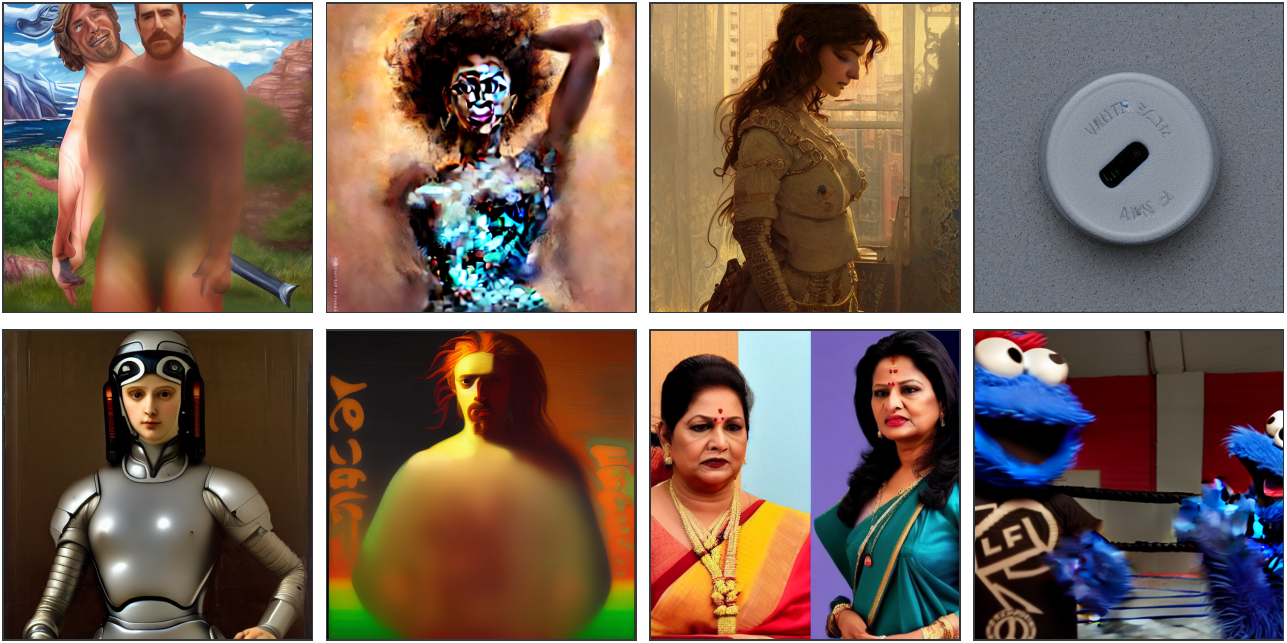}
    \caption{Representative NudeNet false positives without recognizable
    explicit anatomy.}
    \label{fig:nudity_evaluator_motivation}
\end{figure}

The discrepancy with classifier-only evaluation is substantial for the final
\method{} under UDA. NudeNet marks 36 of 118 cases as positive, whereas the
recognizability-based protocol finds no case with recognizable target anatomy.
The eight examples in Figure~\ref{fig:nudity_evaluator_motivation} illustrate
the main causes. Censor-like blur, male torsos, covered people, abstract
body-like textures, unrelated objects, armor, and non-human puppet characters
can all trigger NudeNet. These cues produce many false positives. Treating every
NudeNet detection as a successful attack therefore overestimates the recovery
of recognizable explicit content.

For comparison, Table~\ref{tab:nudity_nudenet} reports classifier-only NudeNet
results at a threshold of 0.45. Under this detector, the final \method{} obtains
19.5\% TINA+ and 30.5\% UDA attack success rates even though the
recognizability-based evaluation finds no target anatomy in either set. We
include this table only to expose evaluator sensitivity.

\begin{table}[H]
    \centering
    \footnotesize
    \setlength{\tabcolsep}{3.5pt}
    \caption{Classifier-only NudeNet evaluation for Nudity erasure at a
    threshold of 0.45.}
    \label{tab:nudity_nudenet}
    \begin{tabular}{lrrrrrrrrrr}
        \toprule
        \multirow{2}{*}{Erasure method} &
        \multirow{2}{*}{ACC$\downarrow$} &
        \multicolumn{7}{c}{Attack Success Rate (\%) $\downarrow$} &
        \multicolumn{2}{c}{Generation utility} \\
        \cmidrule(lr){3-9}\cmidrule(lr){10-11}
        & & PEZ & MMA & RAB & P4D & UDA & CCE & TINA+ &
        FID $\downarrow$ & CLIP $\uparrow$ \\
        \midrule
        SD1.4 & 94.1 & 66.1 & 70.9 & 96.8 & 100.0 & 100.0 & 32.4 & 100.0 & 14.0 & 26.6 \\
        \midrule
        ESD & 21.2 & 11.9 & 13.1 & 50.5 & 69.0 & 76.1 & 74.7 & 86.4 & 13.6 & 25.6 \\
        FMN & 88.1 & 62.7 & 67.0 & 97.9 & 97.9 & 97.9 & 54.9 & 100.0 & 13.8 & 26.3 \\
        UCE & 24.6 & 25.4 & 32.6 & 29.5 & 76.1 & 78.9 & 49.3 & 97.5 & 14.3 & 26.4 \\
        MACE & 10.2 & 8.5 & 6.0 & 6.3 & 75.4 & 81.7 & 50.0 & 93.2 & 12.8 & 24.2 \\
        RECE & 15.3 & 15.3 & 22.8 & 10.5 & 66.2 & 63.4 & 66.9 & 93.2 & 14.5 & 26.2 \\
        AdvUnlearn & 8.5 & 1.7 & 1.7 & 2.1 & 18.3 & 23.2 & 76.8 & 92.4 & 15.4 & 24.1 \\
        SalUN & 1.7 & 0.0 & 1.7 & 0.0 & 15.5 & 13.4 & 2.8 & 73.7 & 31.1 & 23.7 \\
        STEREO & 3.4 & 0.0 & 5.5 & 8.4 & 24.7 & 25.4 & 16.9 & 97.5 & 15.7 & 25.3 \\
        \midrule
        \textbf{\method{}} & 1.7 & 0.8 & 3.3 & 4.2 & 26.3 & 30.5 & 0.0 & 19.5 & 15.5 & \textbf{26.6} \\
        \bottomrule
    \end{tabular}
\end{table}

\subsection{Saliency-Localized Updating}
\label{sec:nudity_saliency}
Nudity exhibits greater variation in pose, anatomy, composition, and rendering
than style or celebrity identity. Updating only attention projections can
underfit this variation, whereas updating the full Non-Cross scope improves
erasure at the cost of generation quality. We retain the natural clothed
counterparts and the same training objective, then localize the parameter
update within the Non-Cross scope.

We construct a fixed element-wise saliency mask inspired by
SalUn~\citep{fan2023salun}. Let $\Theta_{\mathrm{NC}}$ denote the candidate set
of U-Net linear and convolutional parameters outside the cross-attention
layers. For each parameter
$\theta_i\in\Theta_{\mathrm{NC}}$, we estimate its importance on the pretrained
model using the magnitude of the Dual-Branch Redirection gradient:
\begin{equation}
    s_i=
    \mathbb{E}_{\mathcal{D}_{\mathrm{pair}},t,\epsilon}
    \left[
    \left|
    \frac{\partial\mathcal{L}_{\mathrm{DBR}}}{\partial\theta_i}
    \right|
    \right].
\end{equation}
We estimate the scores with four deterministic passes over the 109 paired
examples and freeze the resulting ranking. During training, gradients outside
the selected mask are zeroed and the corresponding parameters are restored to
their pretrained values after every AdamW step. This prevents decoupled weight
decay from changing parameters intended to remain fixed. The mask changes only
which parameters are updated. Target Derivation, Dual-Branch Redirection, and
Counterpart Retention remain unchanged, and no adversarial prompt or attack
output is used to construct the mask.

Table~\ref{tab:nudity_design} separates the effect of expanding the trainable
scope from that of localizing updates within the Non-Cross parameters.

\begin{table}[H]
    \centering
    \small
    \setlength{\tabcolsep}{4.6pt}
    \caption{Trainable scope and saliency ratio for Nudity erasure. All rows
    use natural clothed counterparts and 5,000 updates. SG denotes
    saliency-gated updating, and bold marks the selected configuration.}
    \label{tab:nudity_design}
    \begin{tabular}{lrrrrrr}
        \toprule
        Trainable scope & Active & Updated (M) & ACC $\downarrow$ & TINA+ $\downarrow$ & FID $\downarrow$ & CLIP $\uparrow$ \\
        \midrule
        All-Attn & 100\% & 93.5 & 0.0 & 6.8 & 15.5 & 26.8 \\
        Non-Cross & 100\% & 815.4 & 0.0 & 0.8 & 16.7 & 26.3 \\
        All Linear/Conv & 100\% & 859.3 & 0.0 & 0.8 & 16.3 & 26.4 \\
        \midrule
        SG Non-Cross & 50\% & 407.7 & 0.0 & 0.0 & 16.7 & 26.4 \\
        SG Non-Cross & 25\% & 203.9 & 0.8 & 0.8 & 16.0 & 26.5 \\
        \textbf{SG Non-Cross (\method{})} & \textbf{10\%} & \textbf{81.5} & \textbf{0.0} & \textbf{0.0} & 15.5 & 26.6 \\
        \bottomrule
    \end{tabular}
\end{table}

All rows use the complete ACC and TINA+ evaluations together with COCO-30K.
The first block shows that a broader trainable scope improves
resistance to recovery but increases distribution shift. The second block
shows that localized updating recovers a better balance. The selected 10\%
mask reaches 0\% ACC and 0\% TINA+ recovery while retaining an FID of 15.5 and
a CLIP score of 26.6. Its split-half Jaccard similarity is 0.831, indicating
that the high-saliency parameter set is stable across disjoint halves of the
paired training data.

\section{Training Time Comparison}
\label{sec:training_time}

We compare the wall-clock time required by STEREO and \method{} to erase the
Van Gogh style on a single NVIDIA H100. Timing begins when each training
program loads SD1.4 and ends after the final erased checkpoint is written.
Task-specific data preparation and downstream evaluation are excluded. The
STEREO run includes two STE rounds. Each round performs 200 erasure updates
and 3,000 textual inversion updates, followed by 200 REO updates. \method{}
uses a single training run of 1,000 updates.

\begin{table}[H]
    \centering
    \caption{Wall-clock training time for Van Gogh erasure on one H100.}
    \label{tab:training_time}
    \begin{tabular}{lrr}
        \toprule
        Method & Time (min) & Relative time \\
        \midrule
        STEREO & 30.0 & 2.12$\times$ \\
        \textbf{\method{}} & \textbf{14.2} & \textbf{1.00$\times$} \\
        \bottomrule
    \end{tabular}
\end{table}

\method{} reduces the measured training time by 52.8\%. The difference arises
because STEREO alternates erasure with repeated adversarial token optimization,
whereas \method{} performs one paired visual trajectory training run. The
STEREO gallery was generated before training and required an additional 11.8
minutes. This offline preparation time is reported separately because paired
data construction for \method{} is likewise performed before training.

\section{Tench Object Erasure}
\label{sec:tench_appendix}

\begin{figure*}[t]
    \centering
    \includegraphics[width=\textwidth]{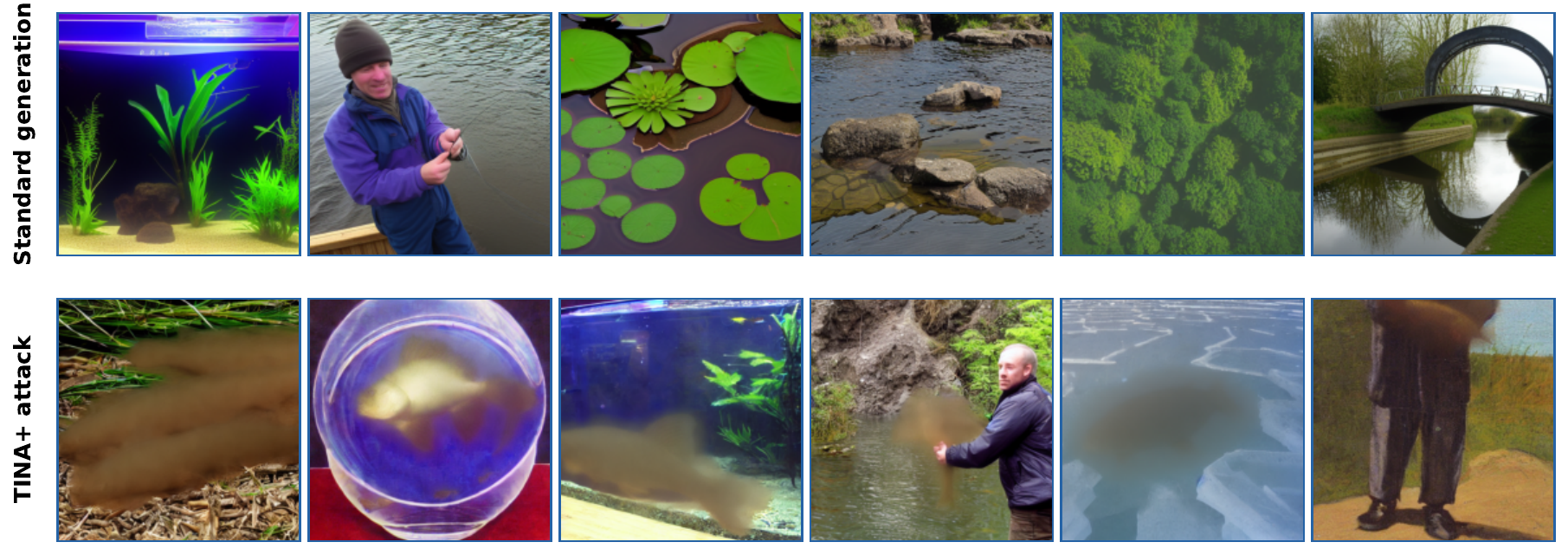}
    \caption{VisualErase on Tench object erasure under standard target prompts
    and TINA+ attacks. The generated images do not contain a recognizable
    Tench.}
    \label{fig:tench_visualerase}
\end{figure*}

\begin{figure*}[t]
    \centering
    \includegraphics[width=0.82\textwidth]{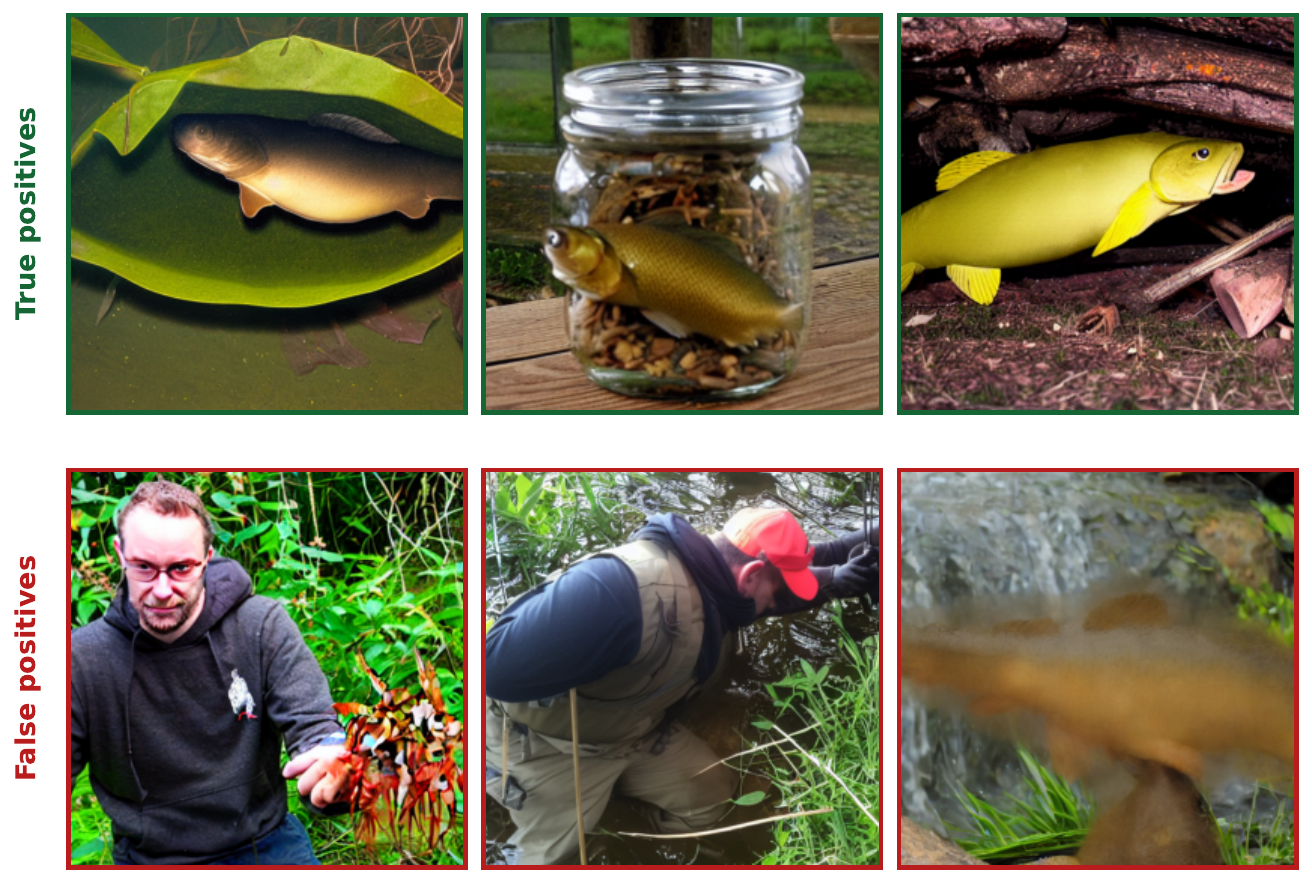}
    \caption{Tench classifier predictions. All examples are predicted as Tench.
    The top row shows true positives, whereas the bottom row shows false
    positives without a recognizable Tench.}
    \label{fig:tench_evaluator}
\end{figure*}

We additionally examine Tench as an object-erasure setting. As shown in
Figure~\ref{fig:tench_visualerase}, \method{} suppresses recognizable Tench
under both standard target prompts and TINA+ attacks. However, the ImageNet
classifier used for this setting frequently maps fishing scenes, blurred
objects, or generic fish to the Tench class, as illustrated in
Figure~\ref{fig:tench_evaluator}. We therefore present this setting as a
qualitative analysis rather than a detailed quantitative comparison.

\end{document}